\pdfoutput=1

\documentclass[11pt]{article}

\usepackage{naacl2021}

\usepackage{times}
\usepackage{latexsym}

\usepackage[T1]{fontenc}

\usepackage[utf8]{inputenc}

\usepackage{microtype}

\usepackage{xspace}
\usepackage{url}
\usepackage{mathtools}

\usepackage{graphicx}
\usepackage{amsmath}
\usepackage{amssymb}
\usepackage{amsthm}
\usepackage{colortbl}
\usepackage{multirow}
\theoremstyle{definition}

\newtheorem{definition}{Definition}[section]
\newtheorem{remark}{Remark}[section]

\newcommand{\stitle}[1]{\vspace{0.6em}\noindent{\bf #1}}
\newcommand{\modelname}{\texttt{BEUrRE}\xspace{}}

\newcommand{\Vol}{\operatorname{Vol}}
\renewcommand{\Box}{\operatorname{Box}}
\newcommand{\cen}{\operatorname{cen}}
\newcommand{\off}{\operatorname{off}}

\newcommand{\boxtext}{\mathrm{Box}}
\newcommand{\boxmin}{\text{m}}
\newcommand{\boxmax}{\text{M}}

\newcommand{\boxspace}{\Omega_{\text{Box}}}
\newcommand{\triple}[3]{\textsl{(#1, #2, #3)}}
\newcommand{\rel}[1]{\textsl{#1}}  
\newcommand{\norm}[1]{\left\lVert#1\right\rVert}
\newcommand{\pbox}{P_{\text{Box}}}
\newcommand{\entbox}[1]{\text{Box}(#1)}

\title{Probabilistic Box Embeddings for Uncertain Knowledge Graph Reasoning}

\author{Xuelu Chen$^{1*}$, Michael Boratko$^{2}$\thanks{\;\;Indicating equal contribution.}\;, Muhao Chen$^{3,4}$\\\textbf{Shib Sankar Dasgupta$^{2}$, Xiang Lorraine Li$^{2}$, Andrew McCallum$^2$}\\
$^1$Department of Computer Science, UCLA\\
$^2$College of Information and Computer Sciences, UMass Amherst\\
$^3$Department of Computer Science, USC\\
$^4$Information Sciences Institute, USC\\
\texttt{shirleychen@cs.ucla.edu}; \texttt{mboratko@iesl.cs.umass.edu};\\ \texttt{muhaoche@usc.edu};  \texttt{\{ssdasgupta,xiangl,mccallum\}@cs.umass.edu}\\ 
}

\date{}

\begin{document}
\maketitle

\begin{abstract}
Knowledge bases often consist of facts which are harvested from a variety of sources, many of which are noisy and some of which conflict, resulting in a level of \emph{uncertainty} for each triple.
Knowledge bases are also often incomplete, prompting the use of embedding methods to generalize from known facts, however 
existing embedding methods only model triple-level uncertainty and reasoning results lack global consistency.
To address these shortcomings, we propose \modelname~\includegraphics[height=1em]{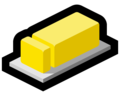}, a novel uncertain knowledge graph embedding method with calibrated probabilistic semantics.
\modelname~models each entity as a \emph{box} (i.e. axis-aligned hyperrectangle),
and relations between two entities as affine transforms on the head and tail entity boxes.
The geometry of the boxes allows for efficient calculation of intersections and volumes, endowing the model with calibrated probabilistic semantics and facilitating the incorporation of relational constraints.
Extensive experiments on two benchmark datasets show that \modelname~consistently outperforms baselines on confidence prediction and fact ranking due to it's probabilistic calibration and ability to capture high-order dependencies among facts.\footnote{Resources and software are available at \url{https://github.com/stasl0217/beurre}}
\end{abstract}

\section{Introduction}\label{sec:intro}
Knowledge graphs (KGs) provide structured representations of facts about real-world entities and relations. 
In addition to deterministic KGs \cite{bollacker2008freebase,lehmann2015dbpedia,mahdisoltani2014yago3}, much recent attention has been paid to uncertain KGs (or UKGs).
UKGs, such as ProBase \cite{wu2012probase}, NELL \cite{mitchell2018never}, and ConceptNet \cite{speer2017conceptnet}, associate each fact (or triple) with a confidence score representing the likelihood of that fact to be true. 
Such uncertain knowledge representations critically capture the uncertain nature of reality, and provide more precise reasoning.
For example, while both 
\triple{Honda}{competeswith}{Toyota} and \triple{Honda}{competeswith}{Chrysler} look somewhat correct, the former fact should have a higher confidence than the latter one, since Honda and Toyota are both Japanese car manufacturers and have highly overlapping customer bases. Similarly, while 
\triple{The Beatles}{genre}{Rock} and \triple{The Beatles}{genre}{Pop} 
are both true, the first one may receive a slightly higher confidence, since the Beatles is generally considered a rock band.
Such confidence information is important when answering questions like \emph{Who is the main competitor of Honda?}, or 
extracting confident knowledge for drug repurposing \cite{sosa2020literature}.

\begin{figure}[t]
    \centering
    \includegraphics[width=\linewidth]{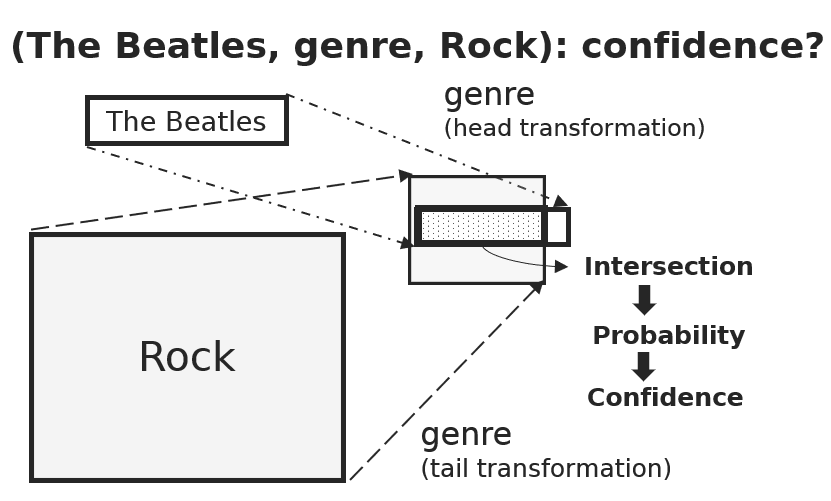}
    \caption{\modelname~models entities as boxes and relations as two affine transforms.} 
    \label{fig:intro_affine}
\end{figure}


To facilitate automated knowledge acquisition for UKGs,
some UKG embedding models 
\cite{chen2019uncertain,kertkeidkachorn2019gtranse} have recently been proposed. Inspired by the works about deterministic KG embeddings \cite{yang2014embedding, bordes2013translating},
existing approaches model entities and relations as points in low-dimensional vector space, measure triple plausibility with vector similarity (eg. distance, dot-product), and map the plausibility to the confidence range of $[0,1]$.
For instance, the representative work UKGE \cite{chen2019uncertain} 
models the triple plausibility in the form of embedding product \cite{yang2014embedding},
and trains the embedding model as a regressor to predict the confidence score.
One interpretation of existing methods is that they model each triple using a binary random variable, where the latent dependency structure between different binary random variables is captured by vector similarities.
Without an explicit dependency structure it is difficult to enforce logical reasoning rules to maintain global consistency.
In order to 
go beyond triple-level uncertainty modeling, we consider each entity as a binary random variable.
However, representing such a probability distribution in an embedding space and reasoning over it is non-trivial.
It is difficult to model marginal and joint probabilities for entities using 
simple geometric objects like vectors.
In order to encode probability distributions in the embedding space,
recent works \cite{lai2017learning, vilnis2018probabilistic, li2019smoothing, dasgupta2020improving} represent random variables as more complex geometric objects,
such as cones and axis-aligned hyperrectangles (\emph{boxes}), and use \emph{volume} as the probability measure.
Inspired by such advances of probability measures in embeddings, we present \modelname~\includegraphics[height=1em]{figures/1f9c8.png} (\textbf{\underline{B}}ox \textbf{\underline{E}}mbedding for \textbf{\underline{U}}nce\textbf{\underline{r}}tain \textbf{\underline{RE}}lational Data)\footnote{``Beurre'' is French for ``butter''.}.
\modelname~represents entities as boxes.
Relations are modeled as two separate affine transforms on the head and tail entity boxes.
Confidence of a triple is modeled by the intersection between the two transformed boxes. 
Fig.~\ref{fig:intro_affine} shows how a fact about the genre of the Beatles is represented under our framework.

Such representation is not only inline with the human perception that entities or concepts have different levels of granularity, but also allows more powerful domain knowledge representation.
UKGE \cite{chen2019uncertain} has demonstrated that introducing domain knowledge about relation properties (e.g. transitivity) can effectively enhance reasoning on UKGs. While UKGE uses Probabilistic Soft Logic (PSL) \cite{bach2017hinge} to reason for unseen facts and adds the extra training samples to training, such a method can lead to error propagation and has limited scope of application when UKG is sparse.
In our work, we propose sufficient conditions for these relation properties to be preserved in the embedding space and directly model the relation properties by regularizing relation-specific transforms based on constraints. 
This technique is more robust to noise and has wide coverage that is not restricted by the scarcity of the existing triples.
Extensive experiments on two benchmark datasets show that \modelname~effectively captures the uncertainty, and consistently outperforms the baseline models on ranking and predicting confidence of unseen facts.

\section{Related Work}

We discuss two lines of related work.

\stitle{UKG Embeddings.}
A UKG assigns a confidence score to each fact. The development of relation extraction and crowdsourcing in recent years enabled the construction of many large-scale uncertain knowledge bases. 
ConceptNet \cite{speer2017conceptnet} is a multilingual KG of commonsense concepts, where triples are assigned with confidence measures reflecting crowdsourcing agreement.
NELL \cite{mitchell2018never} collects facts from web pages with an active-learnable information extraction model, and measures their confidence scores by semi-supervised learning with the Expectation-Maximum (EM) algorithm.
Probase \cite{wu2012probase} is a general probabilistic taxonomy obtained from syntactic extraction.
Aforementioned UKGs have supported numerous knowledge-driven applications, such as short text understanding \cite{wang2016understanding} and literature-based drug repurposing \cite{sosa2020literature}.

Recently, a few UKG embedding methods have been proposed, which seek to facilitate automated knowledge acquisition for UKGs.
UKGE \cite{chen2019uncertain} is the first work of this kind, which models triple plausibility as product of embedding vectors \cite{yang2014embedding}, and maps the  plausibility to the confidence score range of $[0,1]$. To further enhance the performance, UKGE incorporates PSL based constraints 
\cite{bach2017hinge} to help enforce the global consistency of predicted knowledge.
UOKGE \cite{boutouhami2019uncertain} jointly encodes the graph structure and the ontology structure to improve the confidence prediction performance,
which however requires an additional ontology of entity types that is not always available to all KGs.
In addition to the above UKG embeddings models, there is also a matrix-factorization-based approach URGE that seeks to embed uncertain graphs \cite{hu2017embedding}. However, URGE only considers the node proximity in the networks. URGE cannot handle multi-relational data and only generates node embeddings. 


\stitle{Geometric Embeddings.}
Developing embedding methods to represent elements using geometric objects with more complex structures than (Euclidean) vectors is an active area of study.  \emph{Poincar{\'e} embeddings}~\citep{NIPS2017_7213} represent entities in hyperbolic space, leveraging the inductive bias of negative curvature to fit hierarchies.
\emph{Order embeddings}~\citep{orderembedding} take a region-based approach, representing nodes of a graph using infinite cones, and using containment between cones to represent edges.
\emph{Hyperbolic entailment cones}~\citep{ganea2018hypercone} combine order embeddings with hyperbolic geometry.
While these methods show various degrees of promise when embedding hierarchies, they do not provide scores between entities that can be interpreted probabilistically, which is particularly useful in our setting.

\citet{lai2017learning} extend order embeddings with a probabilistic interpretation by integrating the volume of the infinite cones under the negative exponential measure, however the rigid structure imposed by the cone representation limits the representational capacity, and the resulting model cannot model negative correlation or disjointness. Introduced by \citet{vilnis2018probabilistic}, \emph{probabilistic box embeddings} represent elements using axis-aligned hyperrectangles (or \emph{boxes}). Box embeddings not only demonstrate improved performance on modeling hierarchies, 
such embeddings also capture probabilistic semantics based on box volumes, and are capable of compactly representing conditional probability distributions. 
A few training improvement methods for box embeddings have been proposed~\citep{li2019smoothing,dasgupta2020improving}, and we make use of the latter, which is termed \emph{GumbelBox} after the distribution used to model endpoints of boxes.

While box embeddings have shown promise in representing hierarchies, our work is the first use of box embeddings to represent entities in multi-relational data. 
\emph{Query2Box}~\citep{ren2020query2box} and \emph{BoxE}~\citep{BoxE} make use of boxes in the loss function of their models, however entities themselves are represented as vectors, and thus these models do not benefit from the probabilistic semantics of box embeddings, which we rely on heavily for modeling UKGs. 
In \cite{joint_hierarchy}, the authors demonstrate the capability of box embeddings to jointly model two hierarchical relations, which is improved upon using a learned transform in \cite{boxtobox}.
Similarly to \citet{ren2020query2box} and \citet{boxtobox}, we also make use of a learned transform for each relation, however we differ from \citet{ren2020query2box} in that entities themselves are boxes, and differ from both in the structure of the learned transform.
\section{Background}

Before we move on to the presented method in this work, we use this section to introduce the background of box embeddings and the addressed task.

\subsection{Uncertain Knowledge Graphs}
A UKG consists of a set of weighted triples $\mathcal{G}=\{(l,s_l)\}$. For each pair $(l, s_l)$, $l=(h, r, t)$ is a triple representing a fact where $h, t \in \mathcal{E}$ (the set of entities) and $r \in \mathcal{R}$ (the set of relations), and $s_l \in [0,1]$ represents the confidence score for this fact to be true. 
Some examples of weighted triples from NELL are
\triple{Honda}{competeswith}{Toyota}: 1.00 and
\triple{Honda}{competeswith}{Chrysler}: 0.94.



\stitle{UKG Reasoning.} 
Given a UKG $\mathcal{G}$, the \emph{uncertain knowledge graph reasoning} task seeks to predict the confidence of an unseen fact $(h,r,t)$. 

\subsection{Probabilistic Box Embeddings}
In this section we give a formal definition of probabilistic box embeddings, as introduced by \citet{vilnis2018probabilistic}. A \emph{box} is an $n$-dimensional hyperrectangle, i.e. a product of intervals
\begin{equation*}
    \prod_{i=1}^d [x_i^\boxmin, x_i^\boxmax], \quad \text{where} \quad x_i^\boxmin < x_i^\boxmax.
\end{equation*}
Given a space $\boxspace \subseteq \mathbb R^n$, we define $\mathcal B (\boxspace)$ to be the set of all boxes in $\boxspace$. Note that $\mathcal B (\boxspace)$ is closed under intersection, and the volume of a box is simply the product of side-lengths. \citet{vilnis2018probabilistic} note that this allows one to interpret box volumes as unnormalized probabilities. This can be formalized as follows.


\begin{definition}
Let $(\boxspace, \mathcal E, \pbox)$ be a probability space, where $\boxspace \subseteq \mathbb R^n$ and $\mathcal B(\boxspace) \subseteq\mathcal E$. Let $\mathcal Y$ be the set of binary random variables $Y$ on $\boxspace$ such that $Y^{-1}(1) \in \mathcal B(\boxspace)$. A \emph{probabilistic box embedding} of a set $S$ is a function $:S \to \mathcal Y$.
We typically denote $f(s) =: Y_s$ and $Y_s^{-1}(1) =: \Box(s)$.
\end{definition}

Essentially, to each element of $S$ we associate a box which, when taken as the support set of a binary random variable, allows us to interpret each element of $S$ as a binary random variable. Using boxes for the support sets allows one to easily calculate marginal and conditional probabilities, for example if we embed the elements $\{\textsc{cat}, \textsc{mammal}\}$ as boxes in $\boxspace = [0,1]^d$ with $P_\boxtext$ as Lebesgue measure, then 
\begin{equation*}
\begin{split}
&P(\textsc{mammal} \mid \textsc{cat})
=\pbox(X_\textsc{mammal} | X_\textsc{cat})\\
&=\frac{\Vol(\Box(\textsc{mammal}) \cap \Box(\textsc{cat}))}{\Vol(\Box(\textsc{cat}))}.
\end{split}
\end{equation*}

\subsection{Gumbel Boxes}
We further give a brief description of the \emph{GumbelBox} method, which we rely on for training our box embeddings \citep{dasgupta2020improving}.

As described thus far, probabilistic box embeddings would struggle to train via gradient descent, as there are many settings of parameters and objectives which have no gradient signal. (For example, if boxes are disjoint but should overlap.) To mitigate this, \citet{dasgupta2020improving} propose a latent noise model, where the min and max coordinates of boxes in each dimension are modeled via Gumbel distributions, that is
\begin{align*}
    \Box(X) & = \prod_{i=1}^d [x_i^\boxmin, x_i^\boxmax] \quad \text{where}\\
    x_i^\boxmin &\sim \operatorname{GumbelMax}(\mu_i^\boxmin, \beta), \\
    x_i^\boxmax &\sim \operatorname{GumbelMin}(\mu_i^\boxmax, \beta).
\end{align*}
$\mu_i^\boxmin$ thereof is the \emph{location} parameter, and $\beta$ is the (global) variance.
The Gumbel distribution was chosen due to its min/max stability, which means that the set of all ``Gumbel boxes'' are closed under intersection. \citet{dasgupta2020improving} go on to provide an approximation of the expected volume of a Gumbel box,
\begin{equation*}
\label{eq:gumbel_expected_value}
\begin{split}
\mathbb E&\left[\Vol(\Box(X))\right] \approx\\
&\prod_{i=1}^d \beta\log\left(1+\exp\left(\tfrac{\mu_i^\boxmax - \mu_i^\boxmin}\beta - 2\gamma\right)\right).
\end{split}
\end{equation*}
A first-order Taylor series approximation yields
\begin{equation*}
    \mathbb E[\pbox(X_\textsc{A} \mid X_\textsc{B})] \approx \frac{\mathbb E [\Vol(\Box(A) \cap \Box(B))]}{\mathbb E [\Vol(\Box(B))]},
\end{equation*}
and \citet{dasgupta2020improving} empirically demonstrate that this approach leads to improved learning when targeting a given conditional probability distribution as the latent noise essentially ensembles over a large collection of boxes which allows the model to escape plateaus in the loss function. We therefore use this method when training box embeddings.

\begin{remark}
While we use Gumbel boxes for training, intuition is often gained by interpreting these boxes as standard hyperrectangles, which is valid as the Gumbel boxes can be seen as a distribution over such rectangles, with the Gumbel variance parameter $\beta$ acting as a global measure of uncertainty. We thus make statements such as $\Box(X) \subseteq \Box(Y)$, which, strictly speaking, are not well-defined for Gumbel boxes. However we can interpret this probabilistically as $P(Y \mid X) = 1$ which coincides with the conventional interpretation when $\beta = 0$.
\end{remark}

\section{Method}
In this section, we present our UKG embedding model \modelname. The proposed model encodes entities as probabilistic boxes and relations as affine transforms. We also discuss the method to incorporate logical constraints into learning.

\subsection{Modeling UKGs with Box Embeddings}\label{sec:box_emb}


\modelname~represents entities as Gumbel boxes, and a relation $r$ acting on these boxes by translation and scaling. Specifically, we parametrize a Gumbel box $\Box(X)$ using a center $\cen(\Box(X)) \in \mathbb R^d$ and offset $\off(\Box(X)) \in \mathbb R^d_+$, where the location parameters are given by
\begin{align*}
    \mu_i^\boxmin &= \cen(\Box(X)) - \off(\Box(X)), \\
    \mu_i^\boxmax &= \cen(\Box(X)) + \off(\Box(X)).
\end{align*}
We consider transformations on Gumbel boxes parametrized by a translation vector $\tau\in \mathbb R^d$ and a scaling vector $\Delta \in \mathbb R^d_+$ such that
\begin{align*}
    \cen(f(\Box(X); \tau, \Delta)) &= \cen(\Box(X)) + \tau,\\
    \off(f(\Box(X); \tau, \Delta)) &= \off(\Box(X)) \circ \Delta,
\end{align*}
where $\circ$ is the Hadamard product. 
We use separate actions for the head and tail entities of a relation, which we denote $f_r$ and $g_r$, and omit the explicit dependence on the learned parameters $\tau$ and $\Delta$.
\begin{remark}
Note that these relations are not an affine transformations of the \emph{space}, $\boxspace$, rather they perform a transformation of a \emph{box}.
These functions form an Abelian group under composition, and furthermore define a transitive, faithful group action on the set of (Gumbel) boxes.
\end{remark}
Given a triple $(h,r,t)$, \modelname~models the confidence score using the (approximate) conditional probability given by
\begin{equation*} \label{eq:conf}
    \phi(h,r,t) = \frac{\mathbb E[\Vol(f_r(\Box(h)) \cap g_r(\Box(t)))]}{\mathbb E [\Vol(g_r(\Box(t)))]}.
\end{equation*}

 We can think of the box $f_r(\Box(h))$ as the support set of a binary random variable representing the concept $h$ in the context of the head position of relation $r$, for example $\Box(\textsc{TheBeatles})$ is a latent representation of the concept of The Beatles, and $f_\textsc{genre}(\Box(\textsc{TheBeatles}))$ represents The Beatles in the context of genre classification as the object to be classified.
\subsection{Logical Constraints}
The sparsity of real-world UKGs makes learning high quality representations difficult.
To address this problem, previous work \cite{chen2019uncertain} introduces domain knowledge about the properties of relations (e.g., transitivity) and uses PSL over first-order logical rules to reason for unseen facts and create extra training samples.
While this technique successfully enhances the performance by incorporating constraints based on relational properties, 
the coverage of such reasoning is still limited by the density of the graph. In UKGE, the confidence score of a triple can be inferred and benefit training only if all triples in the rule premise are already present in the KG. This leads to a limited scope of application, particularly when the graph is sparse. 

In our work, we propose sufficient conditions for these relation properties to be preserved in the embedding space and directly incorporating the relational constraints by regularizing 
relation-specific transforms.
Compared to previous work, our approach is more robust to noise since it does not hardcode inferred confidence for unseen triples, and it has wide coverage that is not restricted by the 
scarcity of the existing triples.

In the following, we discuss the incorporation of two logical constraints --- transitivity and composition --- in the learning process.
We use capital letters $A, B, C$ to represent universally quantified entities from UKG and use $\Phi$ to denote a set of boxes sampled from $\mathcal B(\boxspace)$.
    

\stitle{Transitivity Constraint.}
A relation $r$ is \emph{transitive} if $(A, r, B) \land (B, r, C) \implies (A, r, C)$.
An example of a transitive relation is \rel{hypernymy}.

\begin{figure}[t]
    \centering
    \includegraphics[width=0.8\linewidth]{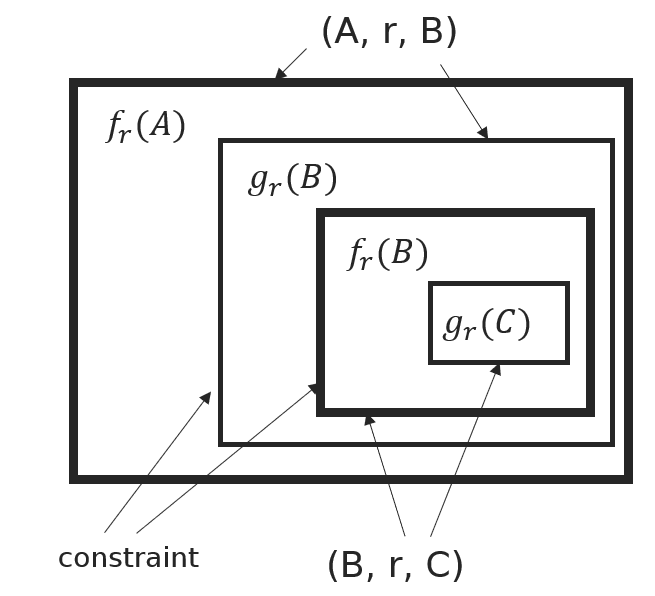}
    \caption{Illustration of how the constraint that $g_r(u)$ contains $f_r(u)$ preserves transitivity of relation $r$ in the embedding space. A triple $(h,r,t)$ is true if and only if $f_r(\entbox{h})$ contains $g_r(\entbox{t}))$. By adding this constraint, $f_r(\entbox{A})$ is guaranteed to contain $g_r(\entbox{C})$ if $(A,r,B)$ and $(B,r,C)$ are true.}
    \label{fig:rule-trans}
\end{figure}

The objective of imposing a transitivity constraint in learning is to preserve this property of the relation in the embedding space, i.e. to ensure that 
$(A, r, C)$ will be predicted true 
if $(A, r, B)$ and $(B, r, C)$ are true.
This objective is fulfilled if $g_r(\entbox{B})$ contains $f_r(\entbox{B})$. An illustration of the box containment relationships is given in Fig \ref{fig:rule-trans}.  
Thus, we constrain $f_r$ and $g_r$ so that $g_r(u)$ contains $f_r(u)$ for any $u \in \boxspace$. We impose the constraint with the following regularization term:
\begin{equation*} 
    L_{\text{tr}}(r) =
        \frac{1}{|\Phi|}
        \sum_{u \in \Phi}
        \norm{
            \pbox(g_r(u) \mid f_r(u)) -1
        }^2.
\end{equation*}

\stitle{Composition Constraint.}
A relation $r_3$ is \emph{composed of} relation $r_1$ and relation $r_2$ if $\quad (A, r_1, B) \land (B, r_2, C)  \implies (A, r_3, C)$.
For example, the relation \rel{atheletePlaysSports} can be composed of relations \rel{atheletePlaysForTeam} and \rel{teamPlaysSports}.

To preserve the relation composition in the embedding space, we constrain that the relation-specific mappings $f_{r_3}$ and $g_{r_3}$ are the \emph{composite mappings} of $f_{r_1}, f_{r_2}$ and $g_{r_1}, g_{r_2}$ respectively:
\begin{equation*}
    f_{r_3} = f_{r_2} \cdot f_{r_1};\;
    g_{r_3} = g_{r_2} \cdot g_{r_1}.
\end{equation*}
where $\cdot$ is the mapping composition operator. 
Thus, for any $u \in \boxspace$, we expect that $f_{r_3}(u)$ is the same as $f_{r_2}( f_{r_1}(u))$ and $g_{r_3}(u)$ is the same as $g_{r_2}(g_{r_1}(u))$.
We accordingly add the following regularization term
\begin{align*} 
\begin{split}
    L_{\text{c}}(r_1, r_2, r_3) 
    =
       &  \frac{1}{|\Phi|}
        \sum_{u \in \Phi}
         f_{r_3}(u) \oplus f_{r_2}( f_{r_1}(u)) \\
        & + g_{r_3}(u) \oplus g_{r_2}(g_{r_1}(u))
\end{split}
\end{align*}
where $\oplus$ is defined as 
\begin{align*}
    \Box_1 \oplus \Box_2 & =  \norm{
            1 - \pbox (\Box_1 \mid \Box_2)
        }^2 \\
        & + \norm{
            1 - \pbox (\Box_2 \mid \Box_1)
        }^2.
\end{align*}

\subsection{Learning Objective}

The learning process of \modelname~optimizes two objectives. The main objective optimizes the loss for a regression task and, simultaneously, a constrained regularization loss enforces the aforementioned constraints.

Let $\mathcal{L}^{+}$ be the set of observed relation facts in training data.
The goal is to minimize the mean squared error (MSE) between the ground truth confidence score $s_l$ and the prediction $\phi(l)$ for each relation $l \in \mathcal{L}^+$. Following UKGE \cite{chen2019uncertain}, we also penalize the predicted confidence scores of facts that are not observed in UKG.
The main learning objective is as follows:
\begin{equation*}
\mathcal{J}_1 = 
\sum_{l \in \mathcal{L}^{+}} |\phi(l)-s_l|^2 
+ \alpha \sum_{l \in \mathcal{L}^{-}} |\phi(l)|^2 .
\end{equation*}
where $\mathcal{L}^{-}$ is a sample set of the facts not observed in UKG, and $\alpha$ is a hyper-parameter to weigh unobserved fact confidence penalization. Similar to previous works, we sample those facts by corrupting the head and the tail for observed facts to generate $\mathcal{L}^{-}$ during training.

In terms of constraints, let $\mathcal{R}_{\text{tr}}$ be the set of transitive relations, $\mathcal{R}_{\text{c}}$ be the set of composite relation groups, and $w_{\text{tr}}$ and $w_{\text{c}}$ be the regularization coefficients.
We add the following regularization to impose our constraints on relations:
\begin{equation*} \label{eq:rule_loss}
\scriptsize{\mathcal{J}_2 = 
    w_{\text{tr}} \sum_{r \in \mathcal{R}_{\text{tr}}} L_{\text{tr}}(r)
    + w_{\text{c}} \!\!\!\sum_{ (r_1, r_2, r_3) \in \mathcal{R}_{\text{c}}} L_{\text{c}}(r_1, r_2, r_3).}
\end{equation*}

Combining both learning objectives, the learning process optimizes the joint loss $J = J_1+J_2$.


\subsection{Inference}

Once \modelname~is trained, the model can easily infer the confidence of a new fact $(h,r,t)$ based on the confidence score function $\phi (h,r,t)$ defined in Section~\ref{sec:box_emb}.
This inference mechanism easily supports other types of reasoning tasks, such as inferring the plausibility of a new fact, and ranking multiple related facts.
The experiments presented in the next section will demonstrate the ability of \modelname~to perform those reasoning tasks.

\section{Experiments}

In this section we present evaluation of our model on two UKG reasoning tasks, i.e. confidence prediction and fact ranking.
More experimentation details are in Appendices.

\subsection{Experiment settings}

\stitle{Datasets.} We follow \citet{chen2019uncertain} and evaluate our models on CN15k and NL27k 
benchmarks, which are subsets of ConceptNet \cite{speer2017conceptnet} and NELL \cite{mitchell2018never}
respectively. 
Table \ref{data_sub} gives the statistics of the datasets.
We use the same split provided by \citet{chen2019uncertain}: 85\% for training, 7\% for validation, and 8\% for testing.
We exclude the dataset PPI5k, the subgraph of the protein-protein interaction (PPI) network STRING \cite{szklarczyk2016string}, 
where the supporting scores of PPI information are indicators based on experimental and literary verification, instead of a probabilistic measure.

\stitle{Logical constraints.}
We report results of both versions of our model with and without logical constraints, denoted as \modelname~(rule+) and \modelname~respectively.
For a fair comparison, we incorporate into \modelname~(rule+) the same set of logical constraints as UKGE \cite{chen2019uncertain}.
Table \ref{table:rule} gives a few examples of the relations on which we impose constraints.

\begin{table}[t]
\centering
\setlength\tabcolsep{2pt}
\begin{tabular}{c|ccccc}
\hline \hline
Dataset & \#Ent. & \#Rel. & \#Rel. Facts & Avg($s$) & Std($s$) \\
\hline
CN15k & 15,000 & 36 & 241,158 & 0.629 & 0.232 \\
NL27k & 27,221 & 404 & 175,412 & 0.797 & 0.242 \\
\hline \hline
\end{tabular}
\caption{Statistics of the datasets. 
\textit{Ent.} and \textit{Rel.} stand for entities and relations. 
Avg($s$) and Std($s$) are the average and standard deviation of confidence.}
\label{data_sub}
\end{table}
\newcommand{\tabincell}[2]{\begin{tabular}{@{}#1@{}}#2\end{tabular}} 
\begin{table}[t]
\centering
\begin{tabular}{c|c|c}
\hline\hline
Dataset & Transitivity & Composition \\
\hline
CN15k
& 
causes
& 
N/A
\\
\hline
NL27k
& 
{\footnotesize locationAtLocation}
& 
\tabincell{c}{ \footnotesize{(atheletePlaysForTeam,}\\ \footnotesize{teamPlaysSport)}\\ \footnotesize{$\rightarrow$ atheletePlaysSport}}
\\
\hline\hline
\end{tabular} \
\caption{Examples of relations with logical constraints.
}
\label{table:rule}
\end{table}

\stitle{Baselines.}
We compare our models with UKG embedding models as well as deterministic KG embedding models.

UKG embedding models include UKGE \cite{chen2019uncertain} and URGE \cite{hu2017embedding}.
While UKGE has multiple versions incorporated with different regression functions, we report the results of the best performing one with the logistic function.
We also include results for both settings with and without constraints, marked as UKGE (rule+) and UKGE in result tables respectively.
URGE was originally designed for probabilistic homogeneous graphs and cannot handle multi-relational graphs, so accordingly we ignore relation information when embedding a UKG.
UOKGE \cite{boutouhami2019uncertain} cannot serve as a baseline because it requires additional ontology information for entities that is not available to these UKGs.

Deterministic KG embedding models TransE \cite{bordes2013translating}, DistMult \cite{yang2014embedding}, ComplEx \cite{trouillon2016complex}, RotatE \cite{sun2018rotate}, and TuckER \cite{balavzevic2019tucker} have demonstrated high performance on reasoning tasks for deterministic KGs, and we also include them as baselines.
These models cannot predict confidence scores for uncertain facts, so we compare our method with them only on the ranking task.
Following \citet{chen2019uncertain}, we only use facts with confidence above the threshold $\tau=0.85$ to train deterministic models.


\stitle{Model configurations.}  
We use Adam \cite{kingma2014adam} as the optimizer and fine-tune the following hyper-parameters by grid search based on the performance on the validation set, i.e. MSE for confidence prediction and normalized Discounted Cumulative Gain (nDCG) for fact ranking. Hyper-parameter search range and the best hyper-parameter configurations are given in Appendix \ref{ap:hyper}. Training terminates with early stopping based on the same metric with a patience of 30 epochs. We repeat each experiment five times and report the average results.

\begin{table}[t]
\centering
\setlength\tabcolsep{2pt}
\begin{tabular}{c|cc|cc}
\hline \hline
Dataset & \multicolumn{2}{c}{CN15k} & \multicolumn{2}{|c}{NL27k} \\
\hline
Metrics & MSE & MAE & MSE & MAE\\
\hline
URGE &10.32  &  22.72  & 7.48  & 11.35 \\
UKGE & 9.02  & 20.05  & 2.67  &7.03\\
\rowcolor[gray]{.9}
\modelname
& 7.80  & 20.03  & 2.37  & 7.12
\\
\hline
UKGE(rule+) & 8.61 & 19.90 & 2.36 &  6.90 \\

\rowcolor[gray]{.9}  
\modelname(rule+)
& \textbf{7.49}  & \textbf{19.88}
& \textbf{2.01}  & \textbf{6.89}
 \\
\hline \hline

\end{tabular}
\caption{Results of fact confidence prediction ($\times 10^{-2}$).}
\label{mse}
\end{table}

\subsection{Confidence Prediction}
This task seeks to predict the confidence of new facts that are unseen to training. 
For each uncertain fact $(l,s_l)$ in the test set, we predict the confidence of $l$ and report the mean squared error (MSE) and mean absolute error (MAE). 
\par

\stitle{Results.}
Results are reported in Table~\ref{mse}.
We compare our models with baselines under the unconstrained and logically constrained (marked with \emph{rule+}) settings respectively.
Under both settings, \modelname~outperforms the baselines in terms of MSE on both datasets. 

Under the unconstrained setting, \modelname~improves MSE of the best baseline UKGE by 0.012 (ca. 14\% relative improvement) on CN15k and 0.003 (ca. 11\% relative improvement) on NL27k.
The enhancement demonstrates that box embeddings can effectively improve reasoning on UKGs.
It is worth noting that even without constraints in learning, \modelname~can still achieve comparable MSE and MAE to the logically constrained UKGE (rule+) on both datasets and even outperforms UKGE (rule+) on CN15k. 
Considering that constraints of relations in CN15k mainly describe transitivity, the aforementioned observation is consistent with the fact that box embeddings are naturally good at capturing transitive relations, as shown in the recent study \cite{vilnis2018probabilistic}. 

With logical constraints, \modelname~(rule+) further enhances the performance of \modelname~and reduces its MSE by 0.0031 (ca. 4\% relative improvement) on CN15k and 0.0036 (ca. 15\% relative improvement) on NL27k.
This is as expected, since logical constraints capture higher-order relations of facts and lead to more globally consistent reasoning.
We also observe that \modelname~(rule+) brings larger gains over \modelname~on NL27k, where we have both transitivity constraints and composition constraints, than on CN15k with only transitivity constraints incorporated. 

In general, with box embeddings, \modelname~effectively improves reasoning on UKGs with better captured fact-wise confidence. 
Furthermore, the results under the logically constrained setting show the effectiveness of improving reasoning with higher-order relations of uncertain facts.

\begin{table}[t]
\centering
\setlength\tabcolsep{1pt}
\begin{tabular}{l|cc}
\hline \hline
Variants & uncons. & rule+ \\
\hline
Metrics & \multicolumn{2}{c}{MSE ($\times 10^{-2}$)} \\
\hline
\modelname & 7.80 & 7.49 \\
\hline
---w/o Gumbel distribution & 8.13 & 8.14 \\
---Single relation-specific transform & 7.81 &  7.60  \\
\hline\hline
\end{tabular}
\caption{Ablation study results on CN15k. \emph{uncons.} represents the unconstrained setting, and \emph{rule+} denotes the logically constrained setting.}
\label{table:ablation}
\end{table}

\stitle{Ablation Study.}
To examine the contribution from Gumbel distribution to model box boundaries and the effectiveness of representing relations as two separate transforms for head and tail boxes, we conduct an ablation study based on CN15k. 
The results for comparison are given in Table \ref{table:ablation}.
First, we resort to a new configuration of \modelname~where we use smoothed boundaries for boxes as in \cite{li2019smoothing} instead of Gumbel boxes. We refer to boxes of this kind as soft boxes.
Under the unconstrained setting, using soft boxes increases MSE by 0.0033 on CN15k (ca. 4\% relative degradation), with even worse performance observed when adding logical constraints.
This confirms the finding by \citet{dasgupta2020improving} that using Gumbel distribution for boundaries greatly improves box embedding training.
Next, to analyze the effect of using separate transforms to represent a relation, we set the tail transform $g_r$ as the identity function. For logical constraint incorporation, we accordingly update the constraint on transitive relation $r$ as $\pbox (u \mid f_r(u))=1, u \in \boxspace$, which requires that $u$ always contains $f_r(u)$, i.e. the translation vector of $f_r$ is always zero and elements of the scaling vector are always less than 1.
Although there is little difference between using one or two transforms under the unconstrained setting, under the logically constrained setting, the constraint is too stringent to be preserved with only one transform.

\begin{table}[t]
\centering
\setlength\tabcolsep{2pt}
\begin{tabular}{c|cc|cc}
\hline \hline
Dataset & \multicolumn{2}{c}{CN15K} & \multicolumn{2}{|c}{NL27k}\\
\hline
Metrics & linear & exp. & linear & exp.  \\
\hline
TransE & 0.601 & 0.591  & 0.730 & 0.722  \\
DistMult & 0.689 & 0.677  & 0.911 & 0.897\\
ComplEx & 0.723 & 0.712  & 0.921 & 0.913\\
RotatE & 0.715 & 0.703 & 0.901  & 0.887 \\
TuckER & 0.736 & 0.724 & 0.877  & 0.870 \\
URGE & 0.572 & 0.570 &0.593 & 0.593 \\
UKGE & 0.769  & 0.768  & 0.933  & 0.929  \\

\rowcolor[gray]{.9} 
\modelname
& 0.796  & 0.795
& 0.942  & 0.942
\\

\hline

UKGE(rule+) & 0.789 & 0.788 & 0.955 & 0.956 \\

\rowcolor[gray]{.9} 
\modelname(rule+)
& \textbf{0.801}  & \textbf{0.803}
& \textbf{0.966}  & \textbf{0.970}
\\
\hline \hline
\end{tabular}
\caption{Mean nDCG for fact ranking. \emph{linear} stands for linear gain, and \emph{exp.} stands for exponential gain.}
\label{rank2}
\end{table}

\stitle{Case study.}
To investigate whether our model can encode meaningful probabilistic semantics, we present a case study about box volumes.  
We examine the objects of the \textsl{atLocation} predicate on CN15k and check which entity boxes have larger volume and cover more entity boxes after the relation transformation. Ideally, geographic entities with larger areas or more frequent mentions should be at the top of the list. When using the \modelname(rule+) model, the top 10 in all entities are \emph{place, town, bed, school, city, home, house, capital, church, camp}, which are general concepts. Among the observed objects of the \emph{atLocation} predicate, the entities that have the least coverage are \emph{Tunisia, Morocco, Algeria, Westminster, Veracruz, Buenos Aires, Emilia-Romagna, Tyrrhenian sea, Kuwait, Serbia}. Those entities are very specific locations. This observation confirms that the box volume effectively represents probabilistic semantics and captures specificity/granularity of concepts, which we believe to be a reason for the performance improvement.

\subsection{Fact Ranking}
Multiple facts can be associated with the same entity. However, those relevant facts may appear with very different plausibility.
Consider the example about Honda Motor Co. in Section~\ref{sec:intro}, where it was mentioned that 
\triple{Honda}{competeswith}{Toyota} should have a higher belief than \triple{Honda}{competeswith}{Chrysler}.
Following this intuition, this task focuses on ranking multiple candidate tail entities for a query $(h, r, \underline{?t})$ in terms of their confidence.

\stitle{Evaluation protocol.} 
Given a query $(h,r,\underline{?t})$, we rank all the entities in the vocabulary as tail entity candidates
and evaluate the ranking performance using the normalized Discounted Cumulative Gain (nDCG) \cite{liu2009learning}.
The gain in retrieving a relevant tail $t_0$ is defined as the ground truth confidence $s_{(h,r,t_0)}$. Same as \citet{chen2019uncertain}, we report two versions of nDCG that use linear gain and exponential gain respectively. The exponential gain puts stronger emphasis on the most relevant results.

\stitle{Results.} 
We report the mean nDCG over the test query set in Table \ref{rank2}.
Although the deterministic models do not explicitly capture the confidence of facts, those models are trained with high-confidence facts and have a certain ability to differentiate high confidence facts from lesser ones.
URGE ignores relation information and yields worse predictions than other models.
UKGE explicitly models uncertainty of facts and is the best performing baseline.

The proposed \modelname~leads to more improvements under both the unconstrained and logically constrained settings.
Under the unconstrained setting, \modelname~offers consistently better performance over UKGE. 
Specifically, on CN15k, \modelname~leads to 0.027 improvement in both linear nDCG and exponential nDCG. 
On NL27k, it offers 0.009 higher linear nDCG and 0.013 higher exponential nDCG. 
Similar to the results on the confidence prediction task, even unconstrained \modelname~is able to outperform the logically constrained UKGE (rule+) on CN15k without incorporating any constraints of relations.
This further confirms the superior expressive power of box embeddings.

In summary, box embeddings improve accuracy and consistency of reasoning and \modelname~delivers better fact ranking performance than baselines.

\section{Conclusion}
This paper presents a novel UKG embedding method with calibrated probabilistic semantics.
Our model \modelname~encodes each entity as a Gumble box representation whose volume represents marginal probability. 
A relation is modeled as two affine transforms on the head and tail entity boxes. We also incorporate logic constraints that capture the high-order dependency of facts and enhance global reasoning consistency.
Extensive experiments show the promising capability of \modelname~on confidence prediction and fact ranking for UKGs.
The results are encouraging and suggest various extensions, including deeper transformation architectures as well as alternative geometries to allow for additional rules to be imposed.
In this context, we are also interested in extending the use of the proposed technologies into more downstream tasks, such as knowledge association \cite{sun2020knowledge} and event hierarchy induction \cite{wang-etal-2020-joint}.
Another direction is to use \modelname~for ontology construction and population, since box embeddings are naturally capable of capturing granularities of concepts.

\section*{Ethical Considerations}
Real-world UKGs often harvest data from open data sources and may include biases. Reasoning over biased UKGs may support or magnify those biases. While not specifically addressed in this work, the ability to inject logical rules could be one way to mitigate bias, and the ability to interpret the learned representation probabilistically allows the investigation of potential learned biases.

All the datasets used in this paper are publicly available and free to download. The model proposed in the paper aims to model uncertainty in knowledge graphs more accurately, and the effectiveness of the proposed model is supported by the empirical experiment results. 

\section*{Acknowledgment}

We appreciate the anonymous reviewers for their insightful comments and suggestions. 

This material is based upon work sponsored by the DARPA MCS program under Contract No. N660011924033 with the United States Office Of Naval Research, and by Air Force Research Laboratory under agreement number FA8750-20-2-10002. 
We also thank our colleagues within IESL at UMass Amherst, for their helpful discussions. Michael, Shib and Xiang were supported in part by the Center for Intelligent Information Retrieval and the Center for Data Science, in part by the IBM Research AI through the AI Horizons Network, in part by the University of Southern California subcontract No. 123875727 under Office of Naval Research prime contract No. N660011924032 and in part by the University of Southern California subcontract no. 89341790 under Defense Advanced Research Projects Agency prime contract No. FA8750-17-C-0106.
The U.S. Government is authorized to reproduce and distribute reprints for Governmental purposes notwithstanding any copyright notation thereon. The views and conclusions contained herein are those of the authors and should not be interpreted as necessarily representing the official policies or endorsements, either expressed or implied, of the U.S. Government.


\bibliographystyle{acl_natbib}
\bibliography{myref}

\newpage
\clearpage
\appendix
\section{Appendices}

\subsection{More Implementation Details}\label{ap:hyper}
Table \ref{tab:hyper} lists hyper-parameter search space for obtaining the set of used numbers. We performed grid search to choose the final setting.
\begin{table}[h]
    \centering
    \footnotesize
    \begin{tabular}{c|c}
    \hline\hline
        Hyper-parameters & Search space  \\
        \hline
        Learning rate $lr$ & \{0.001, 0.0001, 0.00001\}\\\
        Embedding dimension $d$ & \{30, 64, 128, 300\} \\
        Batch size $b$ & \{256, 512, 1024, 2048, 4096\} \\
        Gumbel box temperature $\beta$ & \{0.1, 0.01, 0.001, 0.0001\} \\
        \tabincell{c}{$L_2$ regularization $\lambda$}
         & \{0.001, 0.01, 0.1, 1\} \\
    \hline\hline
    \end{tabular}
    \caption{Search Space for hyper-parameters}
    \label{tab:hyper}
\end{table}

The best hyper-parameter combinations for confidence prediction are 
$\{lr=0.0001, b=1024, d=64, \beta=0.01\}$, $b=2048 $ for CN15k and $b=4096$ for NL27k. $L_2$ regularization is $1$ for box sizes in logarithm scale and $0.001$ for other parameters.
For fact ranking they are $\{lr=0.0001, d=300, b=4096, \lambda=0.00001\}$, $\beta=0.001$ for CN15k and $\beta=0.0001$ for NL27k. 
The number of negative samples is fixed as $30$.
Rule weights are empirically set as $w_{tr}=w_{cp}=0.1$. 

Table \ref{tab:machine} lists the hardware specifications of the machine where we train and evaluate all models. On this machine, training \modelname~for the confidence prediction task takes around 1-1.5 hours.  Training \modelname~for the ranking task takes around 1-2 hours for CN15k and 3 hours for NL27k. 
For the reported model, on CN15k, \modelname~has around 2M parameters for confidence prediction and 9M parameters for ranking. On NL27k, \modelname~has 9M parameters for confidence prediction and 17M for ranking.
\begin{table}[h]
    \centering
    \footnotesize
    \begin{tabular}{c|c}
    \hline\hline
        Hardware & Specification \\
        \hline
        CPU &  $\text{Intel}^{\circledR}$ $\text{Xeon}^{\circledR}$ E5-2650 v4 12-core\\
        GPU & $\text{NVIDIA}^{\circledR}$ GP102 TITAN Xp (12GB)\\
        RAM & 256GB \\
    \hline\hline
    \end{tabular}
    \caption{Hardware specifications of the used machine}
    \label{tab:machine}
\end{table}




\end{document}